\documentclass[10pt,twocolumn,letterpaper]{article}

\usepackage{cvpr}
\usepackage{times}
\usepackage{epsfig}
\usepackage{graphicx}
\usepackage{amsmath}
\usepackage{amssymb}
\usepackage{comment}
\usepackage{multirow}
\usepackage{xcolor}

\usepackage[pagebackref=true,breaklinks=true,letterpaper=true,colorlinks,bookmarks=false]{hyperref}

\cvprfinalcopy 

\definecolor{lightgray}{RGB}{220,220,220}
\definecolor{darkblue}{RGB}{0,0,127}
\definecolor{darkgreen}{RGB}{0,127,0}
\definecolor{darkred}{RGB}{200,0,0}

\ifcvprfinal\fi

\usepackage[affil-it]{authblk}

\usepackage{enumitem}

\title{The 5th AI City Challenge}

\begin{document}

\pagenumbering{gobble}

\author{
Milind Naphade$^1$ \hspace{0.9cm}
Shuo Wang$^1$ \hspace{0.9cm}
David C. Anastasiu$^2$ \hspace{0.9cm}
Zheng Tang$^1$ \\
Ming-Ching Chang$^3$ \hspace{0.9cm}
Xiaodong Yang$^{10}$ \hspace{0.9cm}
Yue Yao$^4$ \hspace{0.9cm}
Liang Zheng$^4$ \\
Pranamesh Chakraborty$^5$ \hspace{0.9cm}
Christian E. Lopez$^6$ \hspace{0.9cm}
Anuj Sharma$^7$ \hspace{0.9cm}
Qi Feng$^8$ \\
Vitaly Ablavsky$^9$ \hspace{0.9cm}
Stan Sclaroff$^8$
} 
\affil{ 
$^1$ NVIDIA Corporation, CA, USA \hspace{0.8cm} 
$^2$ Santa Clara University, CA, USA \\ 
$^3$ University at Albany, SUNY, NY, USA \hspace{0.8cm} 
$^4$ Australian National University, Australia \\
$^5$ Indian Institute of Technology Kanpur, India \hspace{0.8cm} 
$^6$ Lafayette College, PA, USA \\ 
$^7$ Iowa State University, IA, USA \hspace{0.8cm}
$^8$ Boston University, MA, USA \\
$^9$ University of Washington, WA, USA \hspace{0.8cm}
$^{10}$ QCraft, CA, USA \\
}

\maketitle

\begin{abstract}
The AI City Challenge was created with two goals in mind: (1) pushing the boundaries of research and development in intelligent video analysis for smarter cities use cases, and (2) assessing tasks where the level of performance is enough to cause real-world adoption. Transportation is a segment ripe for such adoption. The fifth AI City Challenge attracted 305 participating teams across 38 countries, who leveraged city-scale real traffic data and high-quality synthetic data to compete in five challenge tracks. Track 1 addressed video-based automatic vehicle counting, where the evaluation being conducted on both algorithmic effectiveness and computational efficiency. Track 2 addressed city-scale vehicle re-identification with augmented synthetic data to substantially increase the training set for the task. Track 3 addressed city-scale multi-target multi-camera vehicle tracking. Track 4 addressed traffic anomaly detection. Track 5 was a new track addressing vehicle retrieval using natural language descriptions. The evaluation system shows a general leader board of all submitted results, and a public leader board of results limited to the contest participation rules, where teams are not allowed to use external data in their work. The public leader board shows results more close to real-world situations where annotated data is limited. Results show the promise of AI in Smarter Transportation. State-of-the-art performance for some tasks shows that these technologies are ready for adoption in real-world systems.  
\end{abstract}

\section{Introduction}

The proliferation of sensors has led to the production of fast and voluminous data, along with the emergence and increasing adoption of 5G technologies. The ability to process such voluminous data at the edge has created unique opportunities for extracting insights using the Internet of Things (IoT) for increased operational efficiencies and improved overall outcomes. Intelligent Transportation Systems (ITS) seem ripe to benefit from the adoption of artificial intelligence (AI) applied at the edge. The AI City Challenge was intended to bridge the gap between real-world city-scale problems in ITS and the cutting edge research and development in intelligent video analytics. The challenge is based on data that reflect common scenarios in city-scale traffic management. It also provides an evaluation platform for algorithms to be compared using common metrics. Throughout the past four years of this challenge, we have developed progressively more complex and relevant tasks~\cite{Naphade17AIC17, Naphade18AIC18, Naphade19AIC19, Naphade20AIC20}. 

The fifth edition of this annual challenge, in conjunction with CVPR 2021, continues to push the envelope of research and development in the context of real-world application in several new ways. First, the challenge has introduced a new track for multi-camera retrieval of vehicle trajectories based on natural language descriptions of the vehicles of interest. To our knowledge, this is the first such challenge that combines computer vision and natural language processing (NLP) for city-scale retrieval implementations needed by the Departments of Transportation (DOTs) for operational deployments. The second change in this edition is the expansion of training and testing sets in several challenge tracks. Finally, the vehicle counting track now requires an online, rather than batch algorithm approach to qualify for winning the challenge. Deployment on an edge IoT device helps bring the advances in this field closer to real-world deployment.

The five tracks of the AI City Challenge 2021 are summarized as follows:

\begin{itemize}[leftmargin=12pt] 

\item \textbf{ Multi-class multi-movement vehicle counting using IoT devices:} Vehicle counting is an essential and pivotal task in various traffic analysis activities. The capability to count vehicles under specific movement patterns or categories from a vision-based system is useful yet challenging. This task counts four-wheel vehicles and freight trucks that follow pre-defined movements from multiple camera scenes with online algorithms which should run efficiently on edge devices. The dataset contains 31 video clips of about 9 hours in total that are captured from 20 unique traffic camera views.

\item \textbf{Vehicle re-identification with real and synthetic training data:} Re-identification (ReID)~\cite{dgnet,dgnetpp} aims to establish identity correspondences across different cameras. Our ReID task is evaluated on an expanded version of the previous dataset, referred to as \textit{CityFlowV2-ReID}, which contains over 85,000 vehicle crops captured by 46 cameras placed in multiple traffic intersections. Some of the images are as small as 1,000 pixels. A synthetic dataset~\cite{Yao19VehicleX, Tang19PAMTRI} along with a simulation engine is provided for teams to form augmented training sets.

\item \textbf{City-scale multi-target multi-camera vehicle tracking:} Teams are asked to perform multi-target multi-camera (MTMC) vehicle tracking, whose evaluation is conducted on an updated version of our dataset, referred to as \textit{CityFlowV2}. The annotations on the training set have been refined to include $\sim$60\% more bounding boxes to align with the labeling standard of the test set. There are in total $313,931$ bounding boxes for $880$ distinct annotated vehicle identities.

\item \textbf{Traffic anomaly detection:} 
In this track, teams are required to detect anomalies in videos such as crashes, stalled vehicles, {\em etc.} The dataset used in this track is obtained from video feeds captured at multiple intersections and highways in Iowa, USA. The training set consists of 100 videos, including 18 anomalies, while the test set consists of 150 videos. Each video is in $800 \times 410$ resolution and around 15 minutes long.

\item \textbf{Natural language-based vehicle retrieval:} This newly added task offers natural language (NL) descriptions for teams to specify corresponding vehicle track queries. Participant teams need to perform vehicle retrieval given single-camera tracks and the NL labels. The performance is evaluated using standard retrieval metrics.

\end{itemize}


Continuing the trend of previous editions, this year's AI City Challenge has attracted strong participation, especially with regards to the number of submissions to the evaluation server. We had a total of 305 participating teams that included more than 700 individual researchers from 234 recognized institutions in 38 countries. There were 194, 235, 232, 201, and 155 participation requests received for the 5 challenge tracks, respectively. Of all requesting teams, 137 registered for an account on the evaluation system, and 21, 51, 35, 15, and 20 teams submitted results to the leader boards of the 5 tracks, respectively. Overall, the teams completed 1,685 successful submissions to the evaluation system across all tracks.

This paper presents a detailed summary of the preparation and results of the fifth AI City Challenge. In the following sections, we describe the challenge setup ($\S$~\ref{sec:challenge:setup}), challenge data preparation ($\S$~\ref{sec:dataset}), evaluation methodology ($\S$~\ref{sec:eval}), analysis of submitted results ($\S$~\ref{sec:results}), and a brief discussion of insights and future trends ($\S$~\ref{sec:conclusion}).

\section{Challenge Setup}
\label{sec:challenge:setup}

The fifth AI City Challenge was set up following a similar format as in previous years. The training and test sets were made available to the participants on January 22, 2021. All challenge track submissions were due on April 9, 2021. Similar to the earlier editions, all candidate teams for awards were requested to submit their code for validation. The performance on the leader boards has to be reproducible without the use of external data.

In the released datasets, private information such as vehicle license plates and human faces have been redacted manually. Detailed descriptions of the challenge tasks are as follows.

\textbf{Track 1: Multi-class multi-movement vehicle counting.} Teams were asked to count four-wheel vehicles and freight trucks that followed pre-defined movements from multiple camera scenes. For example, teams performed vehicle counting separately for left-turning, right-turning, and through traffic near a given intersection. This helps traffic engineers understand the traffic demand and freight ratio on individual corridors. Such knowledge can be used to design better intersection signal timing plans and the consideration of traffic congestion mitigation strategies when necessary. To maximize the practical value of the challenge outcome, both vehicle counting effectiveness and the program execution efficiency contributed to the final score evaluation. Additionally, to mimic the performance of in-road hardware sensor-based counting systems, methods were required to run online in real-time. While any system could be used to generate solutions to the problem for general submissions, the final evaluation of the top methods will be executed on an IoT device. The team with the highest combined efficiency and effectiveness score will win this track.

\textbf{Track 2: Vehicle ReID with real and synthetic training data.} Teams were requested to perform vehicle ReID based on vehicle crops from multiple cameras placed at several road intersections. This helps traffic engineers understand journey times along entire corridors. Similar to the previous edition of the challenge, the training set was composed of both real and synthetic data. The usage of synthetic data was encouraged as the simulation engine was provided to create large-scale training sets. The team with the highest accuracy in identifying vehicles that appeared in different cameras will be declared the winner of this track.

\textbf{Track 3: City-scale MTMC vehicle tracking.} Teams were asked to track vehicles across multiple cameras at a single intersection and across multiple intersections spreading out in a mid-size city. Results can be used by traffic engineers to understand traffic conditions at a city-wide scale. The team with the highest accuracy in tracking vehicles that appear in multiple cameras will be declared as the winner. In the event that multiple teams perform equally well in this track, the algorithm needing the least amount of manual supervision will be chosen as the winner.

\textbf{Track 4: Traffic anomaly detection.}  
Based on more than 62 hours of videos collected from different camera views at multiple freeways by the DOT of Iowa, each team was asked to submit a list of at most 100 detected anomalies. The anomalies included single and multiple vehicle crashes and stalled vehicles. Regular congestion was not considered as an anomaly. The team with the highest average precision and the most accurate anomaly starting time prediction in the submitted events will become the winner of this track.

\textbf{Track 5: NL based vehicle retrieval.} In this new challenge track, teams were asked to perform vehicle retrieval given single-view tracks and corresponding NL descriptions of the targets. The performance of the retrieval task was evaluated using the standard metrics of retrieval tasks (\eg, Mean Reciprocal Rank (MRR), Recall@N, \etc.), while ambiguities caused by similar vehicle types, colors, and motion types were considered as well. The NL based vehicle retrieval task offered unique challenges versus action recognition tasks and content-based image retrieval tasks. In particular, different from prior content-based image retrieval systems~\cite{guo2018dialog,hu2016natural,mao2016generation}, retrieval models for this task needed to consider both the relation contexts between vehicle tracks and the motion within each track. While traditional action recognition by NL description~\cite{anne2017localizing} localizes a moment within a video, the NL based vehicle retrieval task requires both temporal and spatial localization within a video.

\section{Datasets}
\label{sec:dataset}

The data used in this challenge were collected from traffic cameras placed in multiple intersections of a mid-size city in USA and the state highways in Iowa. Video feeds have been synchronized manually and the GPS information for some cameras were made available for researchers to leverage the spatio-temporal information. The majority of these video clips are of high resolution (1080p) at 10 frames per second. We have addressed the privacy issues by carefully redacting all vehicle license plates and human faces. In addition to the datasets used in the previous editions of the challenge, a new NL based vehicle retrieval dataset was added for a separate challenge track this year.

Specifically, the following datasets were provided for the challenge this year: 
(1) \textit{CityFlowV2}~\cite{Tang19CityFlow, Naphade19AIC19, Naphade20AIC20} for Track 2 ReID and Track 3 MTMC tracking, (2) \textit{VehicleX}~\cite{Yao19VehicleX, Tang19PAMTRI} for Track 2 ReID, (3) Iowa DOT dataset~\cite{Naphade18AIC18} for Track 1 vehicle counting and Track 4 anomaly event detection, and (4) \textit{CityFlow-NL}~\cite{feng2021cityflownl} for Track 5 NL based vehicle retrieval.

\subsection{The {\bf \textit{CityFlowV2}} dataset}

The \textit{CityFlow} benchmark~\cite{Tang19CityFlow, Naphade19AIC19} was first introduced in the third AI City Challenge in 2019. To the best of our knowledge, it was the first benchmark to address MTMC vehicle tracking in a city scale. A subset of image crops was also created for the task of vehicle ReID. However, there were several issues with this initial release. (1) Many annotations were labeled not properly or missing especially for small-sized objects. (2) The training set was too small compared with the test set, and a validation set was needed. (3) The leading teams in previous years have saturated the performance on the ReID test set.

To continue to challenge the participants, we have upgraded the benchmark in multiple ways this year, and the new version is referred to as \textit{CityFlowV2}. First, we manually refined the annotations of the dataset, especially for the training set, to correct mislabeled objects and include bounding boxes that are as small as $1,000$ pixels. Besides, a new test set containing 6 cameras at multiple intersections on a city highway was introduced in the fourth AI City Challenge. The distance between the two furthest cameras was $4$ km. Moreover, the original test set was adapted to be the validation set for teams to better analyze and improve their models. The number of total bounding boxes has thus grown from $229,680$ to $313,931$, whereas the distinct vehicle identities also increased from $666$ to $880$. Finally, we re-sampled the ReID set for Track 2 with small-sized bounding boxes included, and now there are $85,058$ images versus $56,277$ in the earlier version.

In summary, \textit{CityFlowV2} consisted of $3.58$ hours ($215.03$ minutes) of video captured by $46$ cameras spanning $16$ intersections. The dataset was divided into $6$ simultaneous scenarios, where $3$ were used for training, $2$ for validation, and the other one for testing. Only vehicles that passed through more than one camera were labeled. In each scenario, the time offset and geographic location of each video were provided so that spatio-temporal knowledge can be utilized. The subset for vehicle ReID, namely \textit{CityFlowV2-ReID}, was split into a training set with $52,717$ images from $440$ identities, and a test set including $31,238$ images from another $440$ identities. An additional $1,103$ images were sampled as queries. We also provided in the package evaluation and visualization tools to facilitate the quantitative and qualitative analysis of the results.

\subsection{The {\bf \textit{VehicleX}} dataset}

\begin{figure}[t]
\centerline{
\includegraphics[width=1.1\linewidth]{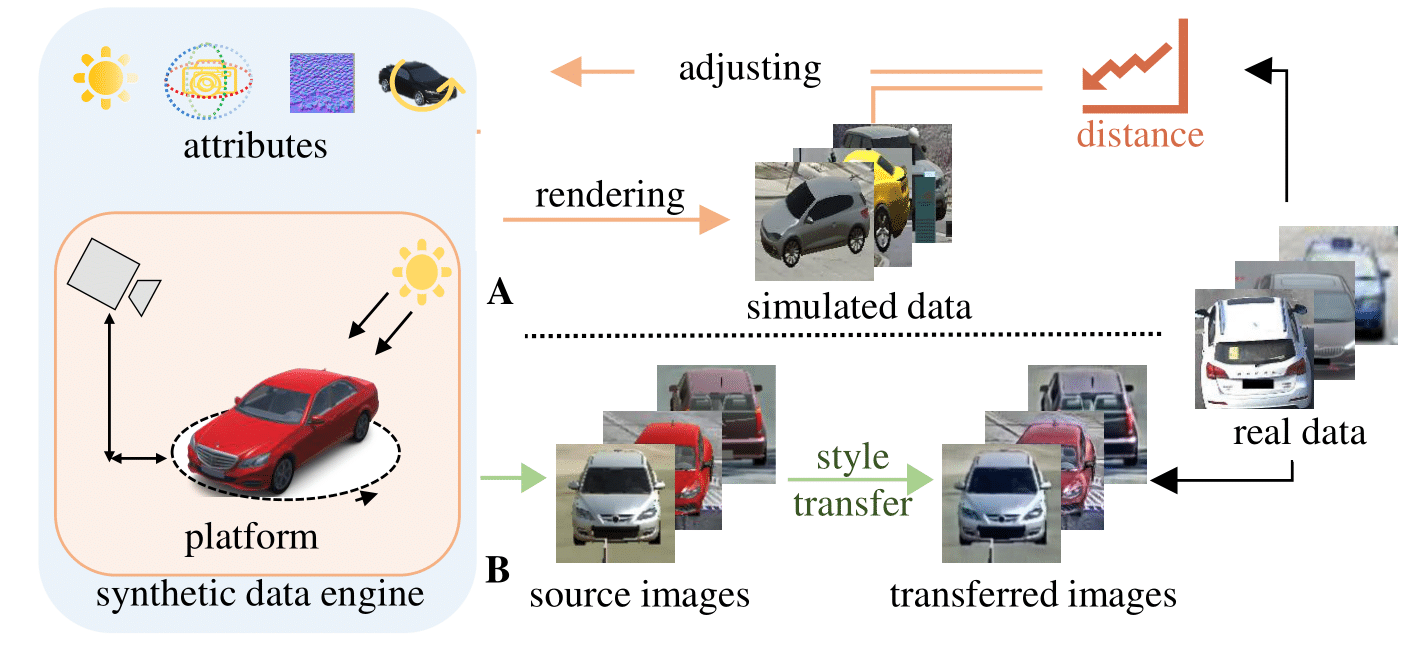}
}
\caption{
{\bf Pipeline for generating \textit{VehicleX} images.} 
With target real data as reference, we used: (A) {\em content-level} domain adaptation which manipulated image contents such as illumination and viewpoint, and (B) {\em appearance (style)-level} domain adaptation which translated image styles. Such simulated images, together with the real ones, were provided to teams for model training.}
\label{fig:vehiclex}
\vspace{-0.4cm}
\end{figure}

The \textit{VehicleX} dataset~\cite{Yao19VehicleX, Tang19PAMTRI} was first introduced in the fourth AI City Challenge in 2020. It has a large number of different types of backbone models and textures that were hand-crafted by professional 3D modelers. To the best of our knowledge, it is currently the largest publicly available 3D vehicle dataset with 11 vehicle types and 272 backbones. Rendered by Unity~\cite{juliani2018unity}, a team can potentially generate an unlimited number of identities and images by editing various attributes. In this year's AI City Challenge, 1,362 identities and more than 190,000 images were generated for joint training with the real-world datasets (\emph{i.e.,} \textit{CityFlowV2-ReID}) to improve the ReID accuracy. We also provided the Unity-Python interface for participant teams, so they could create more synthetic data if needed. They were enabled to generate new identities using a different color on backbones or generate more images with various orientations, camera parameters, and lighting settings. With these attributes, participants can perform multi-task learning, which would improve the ReID accuracy~\cite{Tang19PAMTRI, lin2019improving}. 

In order to minimize the domain gap between the synthetic data and real-world data, a two-level domain adaptation method was performed as shown in Fig.~\ref{fig:vehiclex}. First, on the {\em content level} via the Unity-Python interface, an {\em attribute descent}~\cite{Yao19VehicleX} approach was incorporated to guide the \textit{VehicleX} data in approximating key attributes in real-world datasets. For example, attributes including vehicle orientation, lighting settings, camera configurations, \etc. in the \textit{VehicleX} engine were successively adjusted according to the Fr\'{e}chet Inception Distance (FID) between synthetic data and real data. Secondly, on the {\em appearance (style) level}, SPGAN~\cite{deng2018image} was used to further adapt the style of the synthetic images to better match that of real-world data. The above two-level adaptation method significantly reduced the domain discrepancy between simulated and real data, thereby making \textit{VehicleX} images visually plausible and similar to the real-world ones.

\subsection{Vehicle counting dataset}

This year, we adopted the same vehicle counting dataset that was introduced in the fourth AI City Challenge~\cite{Naphade20AIC20}. The vehicle counting data set contains $31$ video clips (about $9$ hours in total) captured from $20$ unique camera views. Some cameras provide multiple video clips to cover different lighting and weather conditions. Videos are $960$p or better, and most have been captured at $10$ frames per second. The ground truth counts for all videos were manually created and cross-validated by multiple annotators. 

\subsection{Iowa DOT anomaly dataset}

This year, we are using an extended anomaly dataset compared to the one used in the fourth AI City Challenge~\cite{Naphade20AIC20}. The Iowa DOT anomaly dataset consists of $100$ video clips in the training set and $150$ videos in the test set, compared to the $100$ videos each in the training and test sets used in 2020. Video clips were recorded at $30$ frames per second at a resolution of $800 \times 410$. Each video clip is approximately $15$ minutes in duration and may include a single or multiple anomalies. If a second anomaly is reported while the first anomaly is still in progress, it is counted as a single anomaly. The traffic anomalies consist of single or multiple vehicle crashes and stalled vehicles. A total of 18 such anomalies present in the training set across $100$ clips.



\subsection{The {\bf \textit{CityFlow-NL}} Dataset}

\begin{figure*}
  \includegraphics[width=1.0\linewidth]{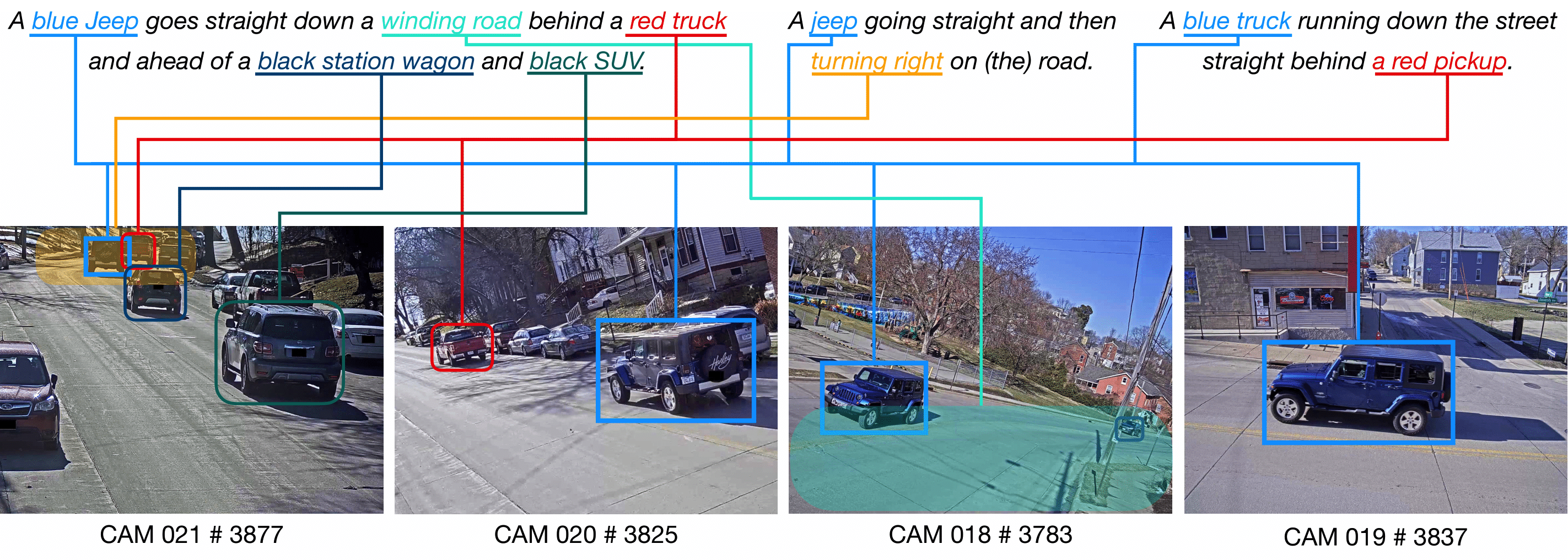}
  \caption{
    The \textbf{\textit{CityFlow-NL} dataset} contains NL descriptions that tend to describe vehicle color/type (\eg, \textit{blue Jeep}), vehicle motion (\eg, \textit{turning right} and \textit{straight}), traffic scene (\eg, \textit{winding road}) and relations with other vehicles (\eg, \textit{red truck}, \textit{black SUV}, \etc).
  }
  \label{fig:cityflow-nl}
  \vspace{-0.2 in}
\end{figure*}

The \textit{CityFlow-NL} benchmark~\cite{feng2021cityflownl} consists of $666$ target vehicles in $3,028$ single-view tracks from $40$ calibrated cameras and $5,289$ unique NL descriptions. For each target, NL descriptions were provided by at least three crowd-sourcing workers, to better capture realistic variations and ambiguities that are expected in the real-world application domains. The NL descriptions describe the vehicle color, vehicle maneuver, traffic scene and relations with other vehicles. Example NL descriptions and targets are shown in Fig.~\ref{fig:cityflow-nl}.

For the NL-based retrieval task, we utilize the \textit{CityFlow-NL} benchmark in a \textit{single-view} setup, although the \textit{CityFlow-NL} can be potentially used for retrieval tasks with multi-view tracks. For each single-view vehicle track, we bundled it with a query consisting of three different NL descriptions for training. During testing, the goal is to retrieve and rank vehicle tracks based on the given NL queries. This variation of the proposed \textit{CityFlow-NL} contains $2,498$ tracks of vehicles with three unique NL descriptions each. Additionally, $530$ unique vehicle tracks together with $530$ query sets (each annotated with three NL descriptions) are curated for testing.

\section{Evaluation Methodology}
\label{sec:eval}

Similar to previous AI City Challenges~\cite{Naphade18AIC18,Naphade19AIC19,Naphade20AIC20}, teams could submit multiple runs (20 for Tracks 2, 3 and 5, and 10 for Tracks 1 and 4) for each track to an \textbf{online evaluation system} that automatically measured the effectiveness of results upon submission. Submissions were limited to five per day, and any submissions that lead to a format or evaluation error did not count against a team's daily or maximum submission totals. During the competition, the evaluation system showed the team's own performance, along with the top-3 best scores on the leader boards (without revealing identifying information of those teams). To discourage excessive fine-tuning to improve performance, the results shown to the teams prior to the end of the challenge were computed on a 50\% subset of the test set for each track. After the challenge submission deadline, the evaluation system revealed the full leader boards with scores computed on the entire test set for each track.

Teams competing for the challenge prizes were not allowed to use external data or manual labeling to fine-tune the performance of their model, and those results were published on the {\bf Public} leader board. Teams using additional external data or manual labeling were allowed to submit to a separate {\bf General} leader board.

\subsection{Track 1 evaluation}

For the first time this year, the multi-class multi-movement vehicle counting task required the creation of \textit{online} algorithms that could generate results in real time. The Track 1 evaluation score (S1) was computed in the same way as in the fourth edition~\cite{Naphade20AIC20}, however ignoring results that would have been produced more than 15 seconds behind real-time playback of the input video. The filtering was done based on self-reported output timestamps which were added to the submission format for the Track 1 challenge. Since efficiency scores reported by teams are not easily normalized, competition prizes will only be awarded based on the scoring obtained when executing the submitted codes from participant teams on the held-out {\em Track 1 Dataset B}. To ensure comparison fairness, Dataset B experiments will be executed on the same device. Also new this year, the target device is an IoT device that could easily be deployed in the field, close-by to the traffic cameras, thereby reducing the need for expensive data transfers to centralized data centers. The target device this year is an NVIDIA Jetson NX development kit.

\subsection{Track 2 evaluation} 
\label{sec:track2:eval}

The Track 2 accuracy evaluation metric was the mean Average Precision (mAP)~\cite{Zheng15Scalable} of the top-$K$ matches, which measured the mean of average precision, \textit{i.e.}, the area under the Precision-Recall curve (AUC) over all the queries. In this track, $K=100$. Our evaluation server also provided other measures, such as the rank-1, rank-5 and rank-10 hit rates, which measured the percentage of the queries that had at least one true positive result ranked within the top 1, 5 or 10 positions, respectively. While these scores were shared with the teams for their own submissions, they were not used in the overall team ranking and were not displayed in the leader boards.

\subsection{Track 3 evaluation}
\label{sec:track1:eval}

The task of Track 3 was detecting and tracking targets across multiple cameras. Baseline detection and single-camera tracking results were provided, and teams were also allowed to use their own methods. Similar to previous years, the IDF1 score~\cite{Ristani16Performance} was used to rank the performance of each team, which measured the ratio of correctly identified detections over the average number of ground-truth and computed detections. The evaluation tool provided with our dataset also computed other evaluation measures adopted by the \textit{MOTChallenge}~\cite{Bernardin2008,Li09Learning}, such as multiple object tracking accuracy (MOTA), multiple object tracking precision (MOTP), mostly tracked targets (MT) and false alarm rate (FAR). However, they were not used for ranking purposes. The measures that were displayed in the evaluation system were IDF1, IDP, IDR, Precision (detection), and Recall (detection).

\subsection{Track 4 evaluation}
\label{sec:track4:eval}

Track 4 performance was measured in the same way as in earlier editions, by combining the detection performance, measured by the $F_1$ score, and detection time error, measured via the normalized root mean square error of the predicted accident times. For full details on the evaluation metric, please see~\cite{Naphade20AIC20}.

\subsection{Track 5 evaluation}

The NL based vehicle retrieval task was evaluated using standard metrics for retrieval tasks~\cite{manning2008introduction}.  We used the Mean Reciprocal Rank (MRR) as the main evaluation metric. Recall@5, Recall@10 and Recall@25 were also evaluated for all models but were not used in the ranking. For a given set $Q$ of queries, the MRR score is computed as
\begin{equation}
    \text{MRR}={\frac{1}{|Q|}}\sum _{i=1}^{|Q|}{\frac {1}{{\text{rank}}_{i}}},
\end{equation}
where ${\text{rank}}_{i}$ refers to the rank position of the first relevant document for the $i$-th query; $|Q|$ is the set size.

\section{Challenge Results}
\label{sec:results}

Tables~\ref{table:1},~\ref{table:2},~\ref{table:3},~\ref{table:4} and~\ref{table:5} summarize the leader boards for Track 1 (turn-counts for signal timing planning), Track 2 (vehicle ReID), Track 3 (city-scale MTMC vehicle tracking), Track 4 (traffic anomaly detection), and Track 5 (NL based vehicle retrieval) challenges, respectively.

\subsection{Summary for the Track 1 challenge}

\begin{table}[t]
\caption{Summary of the Track 1 leader board.}
\label{table:1}
\centering
\footnotesize
\begin{tabular}{|c|c|c|c|}
\hline
Rank & Team ID & Team and paper & Score \\
\hline\hline
1 & 37 & Baidu-SYSU~\cite{BaiduSYSU21Count} & 0.9467 \\
\hline
2 & 5 & HCMIU~\cite{HCMIU21Count} & 0.9459 \\
\hline
3 & 8 & SKKU~\cite{SKKU21Count} & 0.9263 \\
\hline
4 & 19 & HCMUTE~\cite{HCMUTE21Count} & 0.9249 \\
\hline
7 & 95 & Vanderbilt~\cite{Vanderbilt21Count} & 0.8576 \\
\hline
8 & 134 & ComeniusU~\cite{ComeniusU21Count} & 0.8449 \\
\hline
\end{tabular}
\vspace{-0.4cm}
\end{table}

Considering the new on-line performance requirement added this year, teams have put in efforts to balance out effectiveness versus efficiency by designing more computationally efficient algorithms. Similar to last year, a step-by-step {\em detection-tracking-counting} (DTC) framework remained the most popular approach among top-performing teams. There were also new designs not following the DTC framework, which emphasized more on improving execution efficiency.

The top 4 teams~\cite{BaiduSYSU21Count,HCMIU21Count,SKKU21Count,HCMUTE21Count} on the public leader board all followed the DTC framework. All four teams employed YOLO-family models (where~\cite{BaiduSYSU21Count} chose PP-YOLO,~\cite{HCMIU21Count} chose YOLOv4-tiny,~\cite{SKKU21Count} chose YOLOv5, and~\cite{HCMUTE21Count} chose scaled YOLOv4) for vehicle detection. This indicates the popularity of the YOLO models in offering good accuracy as well as computational efficiency. In the tracking step, all four teams adopted the popular SORT tracking strategy with key steps including linear motion prediction, feature extraction, data association, and Kalman filter updates. To accommodate on-line execution requirements, teams either directly used bounding box IOU for data association~\cite{SKKU21Count} or a combination of simpler features including color histogram, motion and shape features, instead of using deep CNN appearance features~\cite{BaiduSYSU21Count,HCMIU21Count,HCMUTE21Count}.

The Baidu-SYSU team~\cite{BaiduSYSU21Count} adopted hand-engineered spacial filters to remove all bounding boxes outside the ROIs before the tracking step, which was simple and effective in suppressing noise. The HCMIU team~\cite{HCMIU21Count} designed a three-fold data association scheme with multiple criteria checkers in a conditional cascade fashion to minimize the computational cost. In the counting step, teams manually drew movement represented tracks and used hand-engineered ROI filters and customized similarity metric to assist vehicle movement assignment. The similarity metric in~\cite{BaiduSYSU21Count} combined Hausdorff distance and the angle between directions. Some teams~\cite{SKKU21Count,HCMUTE21Count} computed Hausdorff distance on each divided sub-segment to count directionality. The HCMIU team~\cite{HCMIU21Count} first applied customized ROIs as filters to make sure all the rest tracks belong to one of the MOIs and then assigned movements by cosine similarity. Additionally, they also implemented thread-level parallelism to boost both robustness and efficiency of their method.

The non-DTC frameworks in~\cite{Vanderbilt21Count,ComeniusU21Count} also showed competitive results.
The CenterTrack object detection and tracking network in~\cite{ComeniusU21Count} generated vehicle bounding boxes as well as location displacement vectors in two consecutive framesin an end-to-end manner (instead of performing detection and tracking in two separated steps).
The localization-based tracking (LBT) in~\cite{Vanderbilt21Count} only detected vehicles on candidate crops from either designated source regions or predicted locations (of already tracked vehicles).
This can effectively avoid repeated object detection on the entire frame.
Both source regions and sink regions were manually defined. 
Their strategy led to reduced computation workloads, as only the vehicles from the source regions were detected and tracked. Vehicle movement was determined based on the combination of the source/sink region of a tracked vehicle. The LBT counting scheme was claimed 52\% faster than the regular DTC framework.

\subsection{Summary for the Track 2 challenge}

\begin{table}[t]
\caption{Summary of the Track 2 leader board.}
\label{table:2}
\centering
\footnotesize
\begin{tabular}{|c|c|c|c|}
\hline
Rank & Team ID & Team and paper & Score \\
\hline\hline
1 & 47 & Alibaba~\cite{Alibaba21ReID} & {\bf 0.7445} \\
\hline
2 & 9 & Baidu-UTS~\cite{BaiduUTS21ReID} & 0.7151 \\
\hline
3 & 7 & SJTU~\cite{SJTU21ReIDMTMCT} & 0.6650 \\
\hline
4 & 35 & Fiberhome~\cite{Fiberhome21ReID} & 0.6555 \\
\hline
9 & 61 & Cybercore~\cite{Cybercore21ReID} & 0.6134 \\
\hline
16 & 54 & UAM~\cite{UAM21ReID} & 0.4900 \\
\hline
21 & 79 & NTU~\cite{NTU21ReIDMTMCT} & 0.4240 \\
\hline
\end{tabular}
\end{table}

Most methods took advantage of the provided synthetic data generator to enhance the real training data. Some teams~\cite{Alibaba21ReID, Cybercore21ReID, NTU21ReIDMTMCT} adopted schemes, \eg, \textit{MixStyle}~\cite{Zhou21MixStyle} and Balanced Cross-Domain Learning, for domain adaptation. For example, the top performing team~\cite{Alibaba21ReID} used unsupervised domain-adaptive (UDA) training to strengthen the robustness. The team from SJTU~\cite{SJTU21ReIDMTMCT} leveraged synthetic data to learn vehicle attributes that better captured appearance features. The other leading team from Baidu-UTS~\cite{BaiduUTS21ReID} proposed a novel part-aware structure-based ReID framework to handle appearance change due to pose and illumination variants. Many teams~\cite{Fiberhome21ReID, UAM21ReID} also extracted video-based features based on query expansion and temporal pooling that suppressed noise in the individual images. Finally, teams reported that post-processing strategies, including re-ranking, image-to-track retrieval, and inter-camera fusion, were useful in improving their methods' effectiveness. 

\subsection{Summary for the Track 3 challenge}

\begin{table}[t]
\caption{Summary of the Track 3 leader board.}
\label{table:3}
\centering
\footnotesize
\begin{tabular}{|c|c|c|c|}
\hline
Rank & Team ID & Team and paper & Score \\
\hline\hline
1 & 75 & Alibaba-UCAS~\cite{AlibabaUCAS21MTMCT} & {\bf 0.8095} \\
\hline
2 & 29 & Baidu~\cite{Baidu21MTMCT} & 0.7787 \\
\hline
3 & 7 & SJTU~\cite{SJTU21ReIDMTMCT} & 0.7651 \\
\hline
4 & 85 & Fraunhofer~\cite{Fraunhofer21MTMCT} & 0.6910 \\
\hline
6 & 27 & Fiberhome~\cite{Fiberhome21MTMCT} & 0.5763 \\
\hline
9 & 79 & NTU~\cite{NTU21ReIDMTMCT} & 0.5458 \\
\hline
10 & 112 & KAIST~\cite{KAIST21MTMCT} & 0.5452 \\
\hline
18 & 123 & NCCU-NYMCTU-UA~\cite{NCCUNYMCTUUA21MTMCT} & 0.1343 \\
\hline
\end{tabular}
\vspace{-0.4cm}
\end{table}

All participant teams followed the typical processing steps for the MTMC vehicle tracking task, including object detection, multi-target single-camera (MTSC) tracking, appearance feature extraction for ReID, and cross-camera tracklet matching. The two best performing teams from Alibaba-UCAS~\cite{AlibabaUCAS21MTMCT} and Baidu~\cite{Baidu21MTMCT} utilized state-of-the-art MTSC tracking schemes, \eg, JDETracker~\cite{Zhang20FairMOT} and TPM~\cite{Peng21TPM}, instead of the provided baseline~\cite{Tang19MOANA, Hsu19MultiCamera} to generate more reliable tracklets. Likewise, the SJTU team~\cite{SJTU21ReIDMTMCT} employed a tracker update strategy to improve the traditional Kalman-filter-based tracking. These teams also leveraged spatial-temporal knowledge and traffic rules to create crossroad zones or entry/exit ports to construct a better distance matrix for tracklet clustering. The Fraunhofer team~\cite{Fraunhofer21MTMCT} proposed an occlusion-aware approach that discarded obstacle-occluded bounding boxes and overlapping tracks for more precise matching. Other teams~\cite{NTU21ReIDMTMCT, KAIST21MTMCT, NCCUNYMCTUUA21MTMCT} also utilized a similar pipeline to extract local trajectories and perform matching using ReID features.

\subsection{Summary for the Track 4 challenge}

\begin{table}[t]
\caption{Summary of the Track 4 leader board.}
\label{table:4}
\centering
\footnotesize
\begin{tabular}{|c|c|c|c|}
\hline
Rank & Team ID & Team and paper & Score \\
\hline\hline
1 & 76 & Baidu-SIAT~\cite{BaiduSIAT21Anomaly} & {\bf 0.9355} \\
\hline
2 & 158 & ByteDance~\cite{ByteDance21Anomaly} & 0.9220 \\
\hline
3 & 92 & WHU~\cite{WHU21Anomaly} & 0.9197 \\
\hline
4 & 90 & USF~\cite{USF21Anomaly} & 0.8597 \\
\hline
5 & 153 & Mizzou-ISU~\cite{MizzouISU21Anomaly} & 0.5686 \\
\hline
\end{tabular}
\vspace{-0.4cm}
\end{table}


The methodologies of the top performing teams in Track 4 of the challenge were based on the basic idea of pre-processing, which involved background modelling, vehicle detection, road mask construction to remove stationary parked vehicles, and abnormal vehicle tracking. 
The dynamic tracking module of the winning team, Baidu-SIAT~\cite{BaiduSIAT21Anomaly}, utilized spatio-temporal status and motion patterns to determine the accurate starting time of the anomalies. Post-processing was performed to further refine the starting time of the traffic anomalies. Their best score was 0.9355, indicating that the problem of traffic anomaly can be solved using current technology. The runner-up, ByteDance~\cite{ByteDance21Anomaly} made use of box-level tracking of the potential spatio-temporal anomalous tubes.
Their method can accurately detect the anomalous time periods using such tubes obtained from background modeling and refinements.
Similarly, the third-place team, WHU~\cite{WHU21Anomaly}, also leveraged box-level and pixel-level tracking to identify anomalies along with a dual modality bi-directional tracing module, which can further refine the time periods.

\subsection{Summary for the Track 5 challenge}

\begin{table}[t]
\caption{Summary of the Track 5 leader board.}
\label{table:5}
\centering
\footnotesize
\begin{tabular}{|c|c|c|c|}
\hline
Rank & Team ID & Team and paper & Score \\
\hline\hline
1 & 132 & Alibaba-UTS-ZJU~\cite{AlibabaUTSZJU21NLRetrieval} & {\bf 0.1869} \\
\hline
2 & 17 & SDU-XidianU-SDJZU~\cite{SDUXidianUSDJZU21NLRetrieval} & 0.1613 \\
\hline
3 & 36 & SUNYKorea~\cite{SUNYKorea21NLRetrieval} & 0.1594 \\
\hline
4 & 20 & Sun Asterisk~\cite{SunAsterisk21NLRetrieval} & 0.1571 \\
\hline
6 & 13 & HCMUS~\cite{HCMUS21NLRetrieval} & 0.1560 \\
\hline
7 & 53 & TUE~\cite{TUE21NLRetrieval} & 0.1548 \\
\hline
8 & 71 & JHU-UMD~\cite{JHUUMD21NLRetrieval} & 0.1364 \\
\hline
10 & 6 & Modulabs-Naver-KookminU~\cite{LeeWL21NLRetrieval} & 0.1195 \\
\hline
11 & 51 & Unimore~\cite{Unimore21NLRetrieval} & 0.1078 \\
\hline
\end{tabular}
\vspace{-0.4cm}
\end{table}

For the NL-based vehicle retrieval task, most teams~\cite{AlibabaUTSZJU21NLRetrieval,SDUXidianUSDJZU21NLRetrieval,SunAsterisk21NLRetrieval,TUE21NLRetrieval,JHUUMD21NLRetrieval,LeeWL21NLRetrieval,Unimore21NLRetrieval} chose to obtain sentence embeddings of the queries, while two teams~\cite{HCMUS21NLRetrieval,SUNYKorea21NLRetrieval} used conventional NLP techniques to process the NL queries. For the cross-modality learning, some teams~\cite{TUE21NLRetrieval,AlibabaUTSZJU21NLRetrieval} utilized the ReID models (approaches from the Track 2 challenge). The adoption of vision models pre-trained on visual ReID data showed improvements from their corresponding baselines. Vehicle motion is an essential part of the NL descriptions in \textit{CityFlow-NL}. Therefore, some teams~\cite{AlibabaUTSZJU21NLRetrieval,LeeWL21NLRetrieval,SUNYKorea21NLRetrieval} developed specific approaches to measure and represent the vehicle motion patterns. 

The best performing model~\cite{AlibabaUTSZJU21NLRetrieval} considered both local (visual) and global representations of the vehicle trajectories to encode vehicle motion. In addition,
NL augmentation via language translation was used to improve the performance of the retrieval. 
The second best performing model~\cite{SDUXidianUSDJZU21NLRetrieval} used GloVe and a custom built gated recurrent unit (GRU)  to measure the similarity between visual crops and the query. 
The method in~\cite{LeeWL21NLRetrieval} retrieved the target vehicle by performing per-frame segmentation, which can be derived as a visual tracker.
However, the performance of this tracker was sub-optimal.~\cite{HCMUS21NLRetrieval} used semantic role labeling techniques on the NL descriptions to rank and retrieve the vehicle tracks.~\cite{SUNYKorea21NLRetrieval} not only considered the vehicle motion but also relations {\em w.r.t.} other vehicles described in the NL queries.
This approach yielded the best performing model not relying on sentence embeddings for similarity measurements.

\section{Conclusion}
\label{sec:conclusion}

The fifth edition of the AI City Challenge continues to attract worldwide research community participation in terms of both quantity and quality. A few observations are noted below.

The main thrust of Track 1 this year was the evaluation of counting methods on edge IoT devices. To this end, teams have put significant efforts in optimizing algorithms as well as implementation pipelines for performance improvement. The detection-tracking-counting (DTC) framework remained the most popular scheme among top-performing teams~\cite{BaiduSYSU21Count,HCMIU21Count,SKKU21Count,HCMUTE21Count}. Within the DTC framework, object tracking was the focus of greater attention. Methods not using deep-appearance-based features proved to be both effective and cost-efficient in the feature extraction and data association. We have also seen innovative designs~\cite{Vanderbilt21Count,ComeniusU21Count} not following the DTC framework and instead emphasizing on execution efficiency and showing competitive results.

In Tracks 2, 3 and 4, we have significantly expanded the datasets, which motivated teams to train more robust models that could be applied to diverse scenarios. In Track 2, we made major improvements to the Unity-Python interface of the synthetic data generator that enabled teams to render vehicle identities of various colors, orientations, camera parameters, light settings, \etc. The top-ranked teams on the leader board leveraged UDA schemes to stabilize training across different domains that resulted in their remarkable performance. In Track 3, a new test set has been added since the fourth edition of the Challenge and the labels of bounding boxes have been largely refined to include more small-sized objects. To tackle this challenging problem, teams utilized state-of-the-art tracking methods to create reliable tracklets and introduced cross-camera matching algorithms based on spatio-temporal information as well as traffic rules and topological structures. As for Track 4, the test set has grown by 150\% compared to the fourth edition in the last year. We have seen teams adopting different types of approaches to detect anomalies, including tracking-based algorithms, background modeling, motion pattern understanding, \etc. 


In Track 5, we proposed a novel challenge for NL based vehicle retrieval. Teams were challenged to apply knowledge across computer vision and NLP to the identification of proper vehicle tracks. Various approaches were introduced by teams to create representative motion features and appearance embeddings. Compared to the other challenge tracks, the performance of the leading teams was far from saturation due to multiple factors. It was difficult to relate the semantic labels to vehicle attributes especially for some that exhibited long tails in the distribution. Moreover, the motion patterns of vehicles required to be described using tracking techniques in 3D space and thus were hard to train directly through NL descriptors. Finally, many vehicles shared similar colors and types, which forced the algorithms to distinguish targets through fine-grained details, a well-known issue for deep learning frameworks. 



Future work for the AI City Challenge will continue pushing the twin objectives of advancing the state of the art and bridging the real-world utility. To this end, while we will continue to increase the dataset sizes, we hope to find forward-thinking DOTs that will provide a platform to deploy some of the most promising approaches emerging out of the AI City Challenge in their operational environments. This new approach will also likely require developing novel evaluation metrics to compare previous status quo baselines with the state-of-the-art AI-based systems developed in the challenge. We believe such a collaboration would make the challenge a truly unique opportunity for ITS applications in the real world and would be of great benefit to the DOTs.

\section{Acknowledgment}

The datasets of the fifth AI City Challenge would not have been possible without significant contributions from the Iowa DOT and an urban traffic agency in the United States. This challenge was also made possible by significant data curation help from the NVIDIA Corporation and academic partners at the Iowa State University, Boston University, Lafayette College, Indian Institute of Technology, Kanpur and Australian National University.


{\small
\bibliographystyle{ieee_fullname}
\bibliography{aicity21, aicity20, aicity19, aicity18, aicity17}
}

\end{document}